\documentclass[letterpaper, 10pt, conference]{ieeeconf}
\IEEEoverridecommandlockouts 
\title{\LARGE \bf
Fusion of stereo and still monocular depth estimates\\ in a self-supervised learning context
}

\usepackage{hyperref}
\usepackage[pdftex]{graphicx}
\usepackage{float}
\usepackage{amsmath}
\usepackage{pgfplots}
\usepackage{tikz}
\usepackage{algorithm}
\usepackage{multirow}
\usepackage{verbatim}

\author{Diogo Martins, Kevin van Hecke,  Guido de Croon}
\begin{document}

\maketitle
\thispagestyle{empty}
\pagestyle{empty}

\begin{abstract}
We study how autonomous robots can learn by themselves to improve their depth estimation capability. In particular, we investigate a self-supervised learning setup in which stereo vision depth estimates serve as targets for a convolutional neural network (CNN) that transforms a single still image to a dense depth map. After training, the stereo and mono estimates are fused with a novel fusion method that preserves high confidence stereo estimates, while leveraging the CNN estimates in the low-confidence regions. The main contribution of the article is that it is shown that the fused estimates lead to a higher performance than the stereo vision estimates alone. Experiments are performed on the KITTI dataset, and on board of a Parrot SLAMDunk, showing that even rather limited CNNs can help provide stereo vision equipped robots with more reliable depth maps for autonomous navigation.
\end{abstract}


\section{Introduction}
\label{sec:introduction}
Accurate 3D information of the environment is essential to several tasks in the field of robotics such as navigation and mapping. Current state-of-the-art technologies for robust depth estimation rely on powerful active sensors like Light Detection And Ranging (LIDAR). Despite the fact that smaller scale solutions as the Microsoft Kinect \citation{Han2013EnhancedCV} exist, they are still too heavy when the available payload and power consumption are  limited, such as on-board of Micro Air Vehicles (MAVs). RGB cameras provide a good alternative, as they can be light, small, and consume little power.

The traditional setup for depth estimation from images consists of a stereo system. Stereo vision has been vastly studied and is considered a reliable method. For instance, NASA's rover Curiosity  was equipped with stereo vision \cite{NASACuriosity} to help detecting potential obstacles in the desired trajectory. However, stereo vision exhibits limited performance in regions with low-texture or with repetitive patterns and when objects appear differently to both views or are partly occluded. Moreover, the resolution of the cameras and the distance between them - baseline - also affect the effective range of accurate depth estimation.
 
Monocular depth estimation is also possible. Multi-view monocular \cite{Engel2014LSDSLAMLD} methods work in  a way similar to stereo vision: single images are captured at different time steps and structures are matched across views. However, opposite to stereo, the baseline is not known, which hampers the process of absolute depth retrieval.  This is a main challenge in this area and typically relies on additional sensors.
 \newpage
 
\usetikzlibrary{automata,positioning,arrows} 
 \begin{figure}[H]
\begin{center}
\begin{tikzpicture}[scale=0.45, every node/.style={scale=0.45}]

\tikzset{
  edge rectangle/.style={
    to path={ rectangle (\tikztotarget)}
  }
}

 \node[draw = black] at (1.9,3.2)  {\includegraphics[width=80 pt, height = 50 pt]{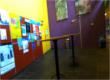}}; 
\node at (1.9,1.8)  { \Large Left stereo image};
\draw[->, thick] (3.4,3.2) -- (4.8, 3.2);

\node[draw = black] at (6.55,3.2)  {\includegraphics[width=80 pt]{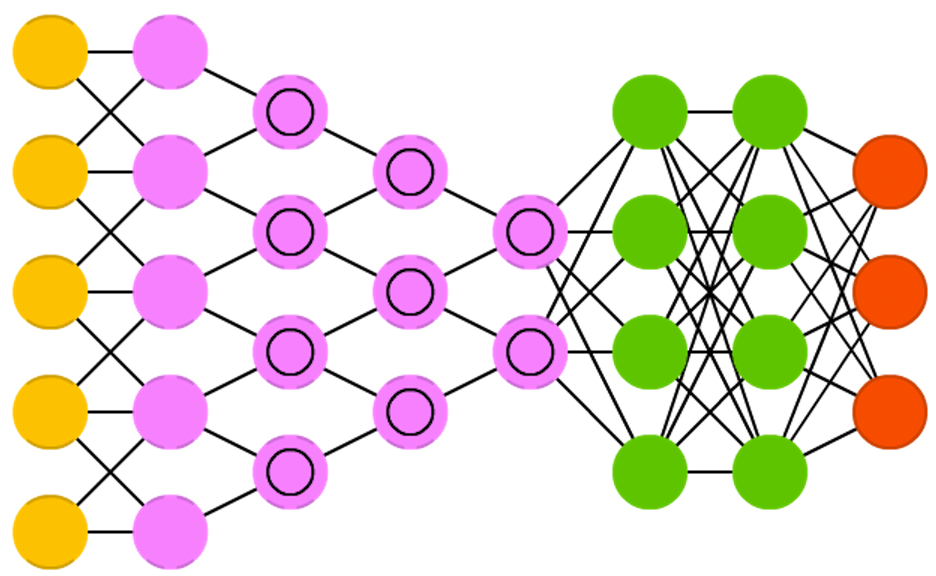}};
\draw[->, thick] (8.1,3.2) -- (9.7, 3.2);
\node[draw = black]  at (11.25,3.2)  {\includegraphics[width=80 pt,  height = 50 pt]{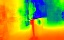}};

\begin{scope}[shift={(0,-3)}]

\draw[->, thick] (3.4,0.45) -- (4.8, 0.45);
 \node[draw=black] at (1.9,0.5)  {\includegraphics[width=80 pt, height = 50 pt]{./left.png}}; 
 \node[draw=black]  at (1.75,0.15)  {\includegraphics[width=80 pt, height = 50 pt]{./left.png}}; 
\node at (1.7,-1.3)  {\Large Stereo pair};
\node[draw = black] at (6.55,0.45)  {\includegraphics[width=80 pt, height = 50 pt]{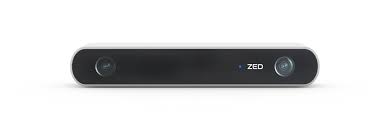}};
\draw[->, thick] (8.1,0.45) -- (9.7, 0.45);
\node[draw = black]  at (11.25,0.45)  {\includegraphics[width=80 pt, height = 50 pt]{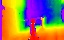}};
\node[color = blue]  at (8.875,-1.3) {\Large Stereo}  ;
\node[color = black]  at (17.2,-1) {\Large Ground Truth} ;
\node[draw = black]  at (17.15,0.45)  {\includegraphics[width=80 pt, height = 50 pt]{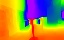}};
\draw[color = blue] (4.8, 1.65) rectangle (12.95,-0.75);
\draw[ thick] (12.8,0.45) -- (14, 0.45);
\end{scope}

\draw[ thick] (12.8,3.2) -- (14, 3.2);
\draw[ ->, thick] (14,3.2) -- (14, 0.59);
\draw[->,  thick] (14,-2.57) -- (14, 0.1);
\node[draw = black, fill = white] at (14,0.35)  {\Large Fusion};
\draw[->,  thick] (14.7,0.35) -- (15.75, 0.35);
\node[draw = black]  at (17.15,0.35)  {\includegraphics[width=80 pt, height = 50 pt]{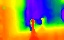}};

\node[color = black]  at (11.15,1.5) {\Large Sparse stereo}  ;
\node[draw = black]  at (11.25,0.35)  {\includegraphics[width=64 pt, height = 40 pt]{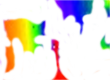}};
\draw[->,  thick] (11.25,-1.55) -- (11.25, -0.525);
\node[draw = black, fill = white] at (8.5,0.35)  {\Large SSL};
\draw[->,  thick] (10.1,0.35) -- (9,0.35);
\draw[->,  thick] (8.5,3.2) -- (8.5,0.6);

\draw[thick] (8.0,0.35) -- (6.57,0.35);
\draw[->,  thick] (6.55,0.335) -- (6.55,1.95);

\draw[color = blue] (4.8, 4.4) rectangle (12.95,2);

\node[color = blue]  at (8.875,4.7) {\Large CNN} ;
\node[color = black]  at (17.15,1.7) {\Large Merged map} ;

\end{tikzpicture}

\end{center}
    \caption{We propose to merge depth estimates from stereo vision with monocular depth estimates from a still image. The robot can learn to estimate depths from still images by using stereo vision depths in a self-supervised learning approach. We show that fusing dense stereo vision and still mono depth gives better results than stereo alone.  \label{fig:mergingOverview} }
\end{figure}

Depth estimation from single still images \cite{Eigen2014DepthMP,Eigen2015PredictingDS,Thesis:Janivecky} - "still-mono" -  provides an alternative to multi-view methods in general. In this case, depth estimation relies on the appearance of the scene and the relationships between its components by means of features, such as texture gradients and color \cite{Saxena2005LearningDF}. The main advantage of still-mono compared to stereo vision is that since only one view is considered, \textit{a priori} there are no limitations in performance imposed by the way objects appear in the field of view or their disposition in the scene. Thus, single mono estimators should not have problems related with very close or very far objects nor when these are partly occluded. As single still-mono depth estimation is less amenable to mathematical analysis than stereo vision, still-mono estimators often rely on learning strategies to infer depths from images \cite{Thesis:Janivecky,Saxena2005LearningDF}  . Thus, feature extraction for depth prediction is done by  minimizing the error on a training set. Consequently, there are no warranties that the model will be able to generalize well to the operational environment, especially if there is a big gap between the operational and training environments.  

A solution to this problem is to have the robot learn depth estimation directly in its own environment. In \cite{garg2016unsupervised} a very elegant method was proposed, making use of the known geometry of the two cameras. In essence, this method trains a deep neural network to predict a disparity map that is then used together with the provided geometrical transformations to reconstruct (or predict) the right image. Follow-up studies have obtained highly accurate depth estimation results in this manner \cite{zhong2017self,godard2017unsupervised}. 

In this article, we explore an alternative path to self-supervised learning of depth estimation in which we assume a robot to be equipped already with a functional stereo vision algorithm. The disparities of this stereo vision algorithm serve as supervised targets for training a deep neural network to estimate disparities from a single still image. Specifically, only sparse disparities in high-confidence image regions are used for the training process. The main contribution of this article is that we show that the \emph{fusion} of the resulting monocular and stereo vision depth estimates gives more accurate results than the stereo vision disparities alone. 
Fig.\ref{fig:mergingOverview} shows an overview of the proposed self-supervised learning setup.

\section{Related work}
\label{sec:relatedWork}
\subsection{Depth estimation from single still images}
\label{subsec:litDepthSingle}
Humans are able to perceive depths with one eye, even if not moving. To this end, we make use of different monocular cues such as occlusion, texture gradients and defocus \cite{FoundVision, Saxena2005LearningDF}. Various computer vision algorithms have been developed over the years to mimic this capability.

The first approaches to monocular depth estimation used vectors of hand-crafted features to  statistically model the scene. These vectors characterize small image patches preserving local structures and include features such as texture energy,  texture  gradients  and haze computed at different scales. Methods such as Markov Random Fields (MRF) have been successfully used  for regression \cite{Saxena2005LearningDF}, while for instance Support Vector Machines (SVMs) have been used to classify each pixel in discrete distance classes \cite{Bipin2015AutonomousNO}. 

In the context of monocular depth estimation,  CNNs are current state-of-the-art \cite{Mancini2017TowardDI, Eigen2015PredictingDS, Liu2016LearningDF}. The use of CNNs forgoes the need of using hand-crafted features. However, large amounts of data are required to ensure full convergence of the solution such that the weight's space is properly explored. 

Despite the fact that different network architectures can be successfully employed, a common approach consists of stacking two or more networks that make depth predictions at different resolutions. One of the networks makes a global, coarse depth prediction that is consecutively refined by the other heaped networks.  These networks will explore local context and incorporate finer-scale details in the global prediction. Different information, such as depth gradients, can also be incorporated \cite{Thesis:Janivecky}. Eigen \textit{et al.} \cite{Eigen2014DepthMP} developed the pioneer study considering CNNs for depth prediction. An architecture consisting of two stacked networks making predictions at different resolutions was used. This architecture was further improved \cite{Eigen2015PredictingDS} by adding one more network for refinement and by performing the tasks of depth estimation, surface normal estimation and semantic labelling jointly. Since this first implementation several other studies have followed using different architectures \cite{Mancini2016FastRM}, posing depth estimation as a classification problem \cite{Cao2016EstimatingDF} or considering a different loss function \cite{Laina2016DeeperDP}. A common ground to these `earlier' deep learning studies is that high quality dense depth maps are used as ground truth during training time. These maps are typically collected using different hardware, such as LIDAR technology or Microsoft Kinect, and are manually processed in order to remove noise or correct wrong depth estimates. 

More recent work has focused on obtaining training data more easily and transferring the learned monocular depth estimation more successfully to the real world. For example, in \cite{mancini2017toward} a Fully Convolutional Network (FCN) is trained to estimate distances in various, visually highly realistic, simulated environments, in which ground-truth distance values are readily available. As mentioned in the introduction, recently, very successful methods have been introduced that learn to estimate distances in a still image by minimizing the reconstruction loss of the right image when estimating disparities in the left image, and viceversa \cite{garg2016unsupervised,zhong2017self,godard2017unsupervised}. Some of these methods are called `unsupervised' by the authors. However, the main learning mechanism is supervised learning and in a robotic context the supervised targets would be generated from the robot's own sensory inputs. Hence, we will discuss these methods under the subsection on self-supervised learning.


\subsection{Fusion of monocular and multi-view depth estimates} 
\label{subsec:fusionSparseDense}

Different approaches have been considered to explore how monocular and multi-view cues (stereo can be posed as a particular case of multi-view where the views are horizontally aligned) can be considered together to increase accuracy of depth estimation. In   \cite{Saxena2007DepthEU} MRFs are used to model depths in an over-segmented image according to an input vector of features. This vector includes (i) monocular cues such as edge filters and texture variations, (ii) the disparity map resultant of stereo matching and (iii) relationships between different small image patches. This model was then trained on a data set collected using a laser scanner. After running the model on the available test set, the conclusion was that the accuracy of depth estimation increases when considering information from monocular and stereo cues jointly. 

A different approach was presented by Facil \textit{et al}. \cite{Fcil2016DeepSA}. Instead of jointly considering monocular and stereo cues, the starting point consists of two finished depth maps: one dense depth map generated by a single view estimator \cite{Eigen2014DepthMP} and a sparse depth map computed using a monocular multi-view method \cite{Engel2014LSDSLAMLD}. The underlying idea is that by combining the reliable structure of the scene given by the CNN's map with the accuracy of selected low-error points from the sparse map it should be possible to generate a final, more accurate depth prediction.  The introduced merging operation is a weighted interpolation of depths over the set of pixels in the multi-view map. The main contribution is an algorithm that improves the depth estimate by merging sparse multi-view with dense mono.  However, there are two remarks which must be made: (i) this study was limited to the fusion of sparse multi-view and dense mono-depth and did not explore the fusion of two dense depth maps and (ii) the CNN was trained and tested in the same environment, which means that  its performance was expected to be good.

We hypothesize that if the CNN was tested in a different environment its performance would be lower, affecting the overall performance of the merging algorithm. Therefore, it is important to incorporate strategies that help reducing the gap between a CNN's training and operational environment. Self-supervised learning is one possible option. 

\subsection{Self-Supervised Learning}
\label{subsec:litSSL}
Self-supervised learning (SSL) is a learning setup in which robots perform supervised learning, where the targets are generated from their own sensors. A typical setup of SSL is one in which the robot uses a trusted primary sensor cue to train a secondary sensor cue. The benefit the robot draws from this, typically lies in the different nature of the sensor cues. For instance, one of the first successful applications of SSL was in the context of the DARPA Grand Challenge, in which  the robot Stanley \cite{Thrun2006StanleyTR} used laser-based technology as supervisory input to train a color model for  terrain classification with a camera. As the camera could see the road beyond the range of the laser scanner, using the stereo system in regions which were not covered by the laser extended the amount of terrain that was properly labeled as drivable or not. Having more information about the terrain ahead helped the team to drive faster and consequently win the challenge. 


Self-supervised learning of monocular depth estimation is a more recent topic \cite{van2015persistent,Hecke2016PersistentSL,lamers2016self}. \cite{Hecke2016PersistentSL} conducted the first study where stereo vision was used as supervisory input to teach a single mono estimator how to predict depths. However, as the focus of the study was more on the behavioral aspects of SSL and all algorithms had to run on a computationally limited robot in space \cite{van2017self}, only the average depth of the scene was learned. Of course, the average depth does not suffice when aiming for more complex navigation behaviors. Hence, in \cite{Thesis:Paquim} a preliminary study was performed on how SSL can be used to train a dense single still  mono estimator. 

Also \cite{garg2016unsupervised,zhong2017self,godard2017unsupervised} learn monocular dense depth estimation, but then by using an image reconstruction loss. Some of these articles use the term `unsupervised learning', as there is no human supervision. Although it is just a matter of semantics, we would put them in the category of `self-supervised learning', since the learning process is supervised and - when used on a robot - the targets come from the robot itself (with the right image values as learning targets when estimating depths in the left image). 


The current study is inspired by \cite{GuidoC}, in which a first study was conducted to understand under which conditions it is beneficial to use `SSL fusion', and in particular, the fusion of a trusted primary sensor cue and a trained secondary sensory cue. Both theoretical and empirical evidence was found that SSL fusion leads to better results when the secondary cue becomes accurate enough. SSL fusion was shown to work on a rather limited real-world case study of height estimation with a sonar and barometer. The goal of this article is not as much to obtain the best depth estimation results known to date, but to present a more complex, real-world case study of SSL fusion. To this end we perform SSL fusion of dense stereo and monocular depth estimates, with the latter learning from sparse stereo targets. The potential merit of this approach lies in showing that the concept of SSL fusion can also generalize to complex real-world cases, where the trusted primary cue is as accurate and reliable as stereo vision.

\section{Methodology overview }
\label{sec:approach}
\label{subsec:sslfusionOverview}

Figure \ref{fig:mergingOverview} illustrates the overall composition of the framework for SSL fusion of stereo and still-mono depth estimates. It can be broken down in four different 'blocks': (i) the stereo estimation, (ii) the still-mono estimation, (iii) fusion of depth estimates and (iv) SSL. 

We expect that the fusion of stereo and monocular vision will give more accurate, dense results than stereo vision alone. This expectation is based on at least two reasons, the first of which is of a geometrical nature (see  fig.\ref{fig:occlusionfilling}). Considering fig. \ref{fig:occlusionfilling}, the two cameras face a brown wall in the background and a black object close-by. Stereo vision cannot provide depth information in the blue regions either because these are not in the field of view of both cameras or because the dark object is occluding them in one of the cameras. If no post-processing was applied, the robot would be "blind" in these areas. However, a single monocular estimator has no problem to provide depth estimates in those regions as it only requires one view. The second reason is while stereo depth estimation relies on triangulation, still monocular depth estimation relies on very different visual cues such as texture density, known object sizes, defocus, etc. Hence,  some problems of stereo vision are in principle not a problem for still-mono: uniform or repetitive textures, very close objects, etc.

 \usetikzlibrary{automata,positioning,arrows} 
 \usetikzlibrary{patterns}
 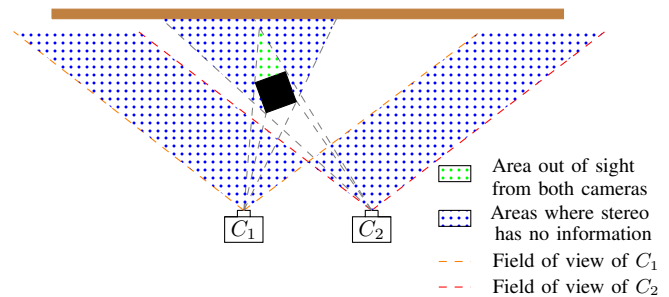
\begin{figure}[H]
\begin{center}
\begin{tikzpicture}[scale=0.85, every node/.style={scale=0.85}]

\node at (-1,-0.3) {$C_1$};
\draw [dashed,red] (1,0) -- (-2.65,2.8);
\draw [dashed,red] (1,0) -- (4.65,2.8);
\draw (-1.1,0)--(-0.9,0)--(-0.9,-0.1) -- (-0.7,-0.1) -- (-0.7,-0.5) -- (-1.3,-0.5) -- (-1.3,-0.1) -- (-1.1,-0.1) -- (-1.1,0);
\draw (-1.3,-0.1) -- (-0.9,-0.1);

\node at (1,-0.3) {$C_2$};
\draw [dashed,orange] (-1,0) -- (-4.6,2.8);
\draw [dashed,orange] (-1,0) -- (2.7,2.8);
\draw (1.1,0)--(0.9,0)--(0.9,-0.1) -- (0.7,-0.1) -- (0.7,-0.5) -- (1.3,-0.5) -- (1.3,-0.1) -- (1.1,-0.1) -- (1.1,0);
\draw (1.3,-0.1) -- (0.9,-0.1);

\draw[fill,brown] (-4,3) -- (-4,3.15) -- (4,3.15) -- (4,3) -- (-4,3); 

\draw [dashed,gray] (1,0) -- (-2.29,3);
\draw [dashed,gray] (-1,0) -- (-0.575,1.85);
\draw [dashed,gray] (-1,0) --  (-0.74390275133,3);

\fill[pattern=dots, pattern color=blue] (-2.29,3) -- (-0.85,3) -- (-0.76,2.85) -- (-0.825,2.05) --  (-0.65,1.5) --  (-2.29,3);

\fill[pattern=dots, pattern color=blue] (-1,0) -- (-4.6,2.8) -- (-2.65,2.8) -- (0,0.75) --  (-1,0);

\fill[pattern=dots, pattern color=blue] (1,0) -- (4.65,2.8) -- (2.7,2.8) -- (0,0.75) --  (1,0);

\draw [dashed,gray] (1,0) --  (-0.84186046511,3);

\draw [dashed,gray] (-1,0) --  (0.45,3);
\draw [dashed,gray] (1,0) --  (-0.375,2);

\begin{scope}[shift={(-0.5,1.85)}]
\draw[fill,black,rotate=20,scale=0.25] (1,1) -- (1,-1) -- (-1,-1) -- (-1,1) -- (1,1); 
  \end{scope}

\fill[pattern=dots, pattern color=blue] (-0.74390275133,3) -- (0.45,3) -- (-0.175,1.7) -- (-0.345,2.1775) --  (-0.76,2.85)--  (-0.74390275133,3);

\begin{scope}[shift = {(-2.2,-2.3)}]
\draw[black,pattern=dots, pattern color=green ] (4.25,3) -- (4.75,3) -- (4.75,2.75) -- (4.25,2.75) -- (4.25,3);
\node at (6.2, 2.975) { \small Area out of sight};
\node at (6.3, 2.65) {\small from  both cameras};

\draw[black,pattern=dots, pattern color=blue ] (4.25,2.3) -- (4.75,2.3) -- (4.75,2.05) -- (4.25,2.05) -- (4.25,2.3);

\node at (6.3, 2.275) { \small {Areas where stereo }};
\node at (6.3, 1.95) {\small { has no information}};

\draw[dashed,orange] (4.25, 1.5) -- (4.75,1.5);
\node at (6.4, 1.5) {\small{Field of view of $C_1$}};

\draw[dashed,red] (4.25, 1.1) -- (4.75,1.1);
\node at (6.4, 1.1) {\small{Field of view of $C_2$}};
\end{scope}

\fill[pattern=dots, pattern color=green]  (-0.76,2.85) -- (-0.345,2.1775) -- (-0.82,2.0)   -- (-0.76,2.85);;

\end{tikzpicture}
\end{center}
    \caption{Example of how stereo vision can benefit from monocular information: the regions where stereo vision is 'blind' can be unveiled by the monocular estimator, as in those areas a still mono estimator has  \textit{a priori} no constraints to make a valid depth prediction. Note that for illustration purposes, the scene and obstacle are quite close to the camera. In large outdoor scenes with obstacles further away, the proportion of occluded areas will be much smaller. \label{fig:occlusionfilling} }
\end{figure}

\subsection{Monocular depth estimation}
\label{subsec:monoEstimate}
The monocular depth estimation is performed with the Fully Convolutional Network (FCN) as used in \cite{mancini2017toward}. The basis of this network is the well known VGG network of \cite{simonyan2014very}, which is pruned of its fully connected layers. Out of the 16 layers of the truncated VGG network, the first 8 were kept fixed, while the others were finetuned for the task of depth estimation. In order to accomodate for this task, in \cite{mancini2017toward} two deconvolutional layers were added to the network that bring the neural representation back to the desired depth map resolution. 

In \cite{mancini2017toward}, the FCN was trained on depth maps obtained from various visually highly realistic simulated environments. In the current study, we will train and fine-tune the same layers, but then using sparse stereo-based disparity measurements as supervised targets. Specifically, we first apply the algorithm of \cite{Hirschmller2008StereoPB} as implemented in OpenCV. Only the disparities at image locations with sufficient vertical contrast are used for training. To this end, we apply a vertical Sobel filter and threshold the output to obtain a binarized map. We use this map as a confidence map for the stereo disparity map.

We use the KITTI data set \cite{Geiger2012CVPR}, employing their provided standard partitioning of training and validation set. The FCN was trained for 1000 epochs. In each epoch 32 images were loaded, and from these images we sampled 100 times a smaller batch of 8 images for training. The loss function used was the mean absolute depth estimation error: $l = \frac{1}{N} \sum_{(x,y) \in C} |Z_{m_{(x,y)}} - Z_{s_{(x,y)}}|$, where $C$ is the set of confident stereo vision estimates. After training, the average absolute loss on the training set is $l = 0.01$.

\subsection{Dense stereo and dense still mono depth fusion}
\label{subsec:denseFusion}
In contrast to \cite{Fcil2016DeepSA}, we propose the fusion of dense stereo vision and dense still mono. There are five main principles behind the fusion operation: (i) as the CNN is better at estimating relative depths \cite{Fcil2016DeepSA}, its output should be scaled to the stereo range, (ii) when a pixel is occluded only monocular estimates are preserved, (iii) when stereo is considered reliable, its estimates are preserved, (iv) when in a region of low stereo confidence and if the relative depth estimates are dissimilar, then the CNN is trusted more, and (v) again when in a region of low stereo confidence but if the relative depth estimates are similar, then the stereo is trusted more.


The scaling is done as follows.
\begin{equation}
Z_{m_{(x,y)}} \leftarrow \textrm{min}(Z_s) + r_s \cdot
\frac{Z_{m_{(x,y)}} - \textrm{min}(Z_m)}{r_m}
\end{equation}
\noindent where $r_m = \textrm{max}(Z_m)-\textrm{min}(Z_m)$ and $r_s = \textrm{max}(Z_s) - \textrm{min}(Z_s)$, and $(x,y)$ is a pixel coordinate in the image. If the stereo output is invalid, as in the case of occluded regions, the depth in the fused map is set to the monocular estimate:
\begin{equation}
Z_{(x',y')} \leftarrow Z_{m_{(x',y')}}
\end{equation}
\noindent where $(x',y')$ is an invalid stereo image coordinate. 

For the remaining coordinates, the depths are fused according to:
\begin{multline}
Z_{(x,y)} 
\leftarrow W_{c_{(x,y)}} \cdot Z_{s_{(x,y)}} + \left ( 1 - W_{c_{(x,y)}} \right ) \cdot \\
\biggl ( W_{s_{(x,y)}} \cdot Z_{s_{(x,y)}} + \left ( 1 - W_{s_{(x,y)}} \right ) \cdot Z_{m_{(x,y)}}
\biggr ) 
\end{multline}
\noindent where $W_{c_{(x,y)}}$ is a weight dependent on the confidence of the stereo map at pixel $(x,y)$,  and $W_{s_{(x,y)}}$ a weight evaluating the ratio between the normalized estimates from the CNN and from the stereo algorithm at pixel $(x,y)$. These weights are defined below.

\begin{figure}[H]
    \centering
    \includegraphics[width = 200 pt, height = 50 pt]{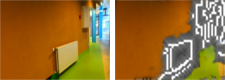}
\caption{\textit{Left:} original left RGB input. \textit{Right:}  stereo confidence map $W_c$ overlaid with original left image. The confidence on the stereo estimate is different than 0 for the white bright pixels, being the whitest ones those where the confidence is maximal. The distribution of high confidence points over the image is condensed especially closer to edges, fading out from there. For instance, there is a negative gradient of brightness (or confidence) going from the borders of the heater to the wall. Moreover, texture-less regions such as the wall are, as expected, classified as low-confidence stereo regions. }
    \label{fig:mergeWeightsSparseCNN}
\end{figure}

Since stereo vision involves finding correspondences in the same image row, it relies on vertical contrasts in the image. Hence, we make $W_{c_{(x,y)}}$ dependent on such contrasts. Specifically, we convolve the image with a vertical Sobel filter and apply a threshold to obtain a binary map. This map is subsequently convolved with a Gaussian blur filter of a relatively large size and renormalized so that the maximal edge value would result in $W_{c_{(x,y)}}=1$. The blurring is performed to capture the fact that pixels close to an edge will likely still be well-matched due to the edge falling into their support region (e.g., matching window in a block matching scheme). Please see fig. \ref{fig:mergeWeightsSparseCNN} for the resulting confidence map $W_c$.


 
 
 
If $W_{c_{(x,y)}} < 1$, the monocular and stereo estimates will be fused together with the help of the weight $W_{s_{(x,y)}}$. In the proposed fusion, more weight will be given to the stereo vision estimate, if $Z_{s_{(x,y)}}$ and $Z_{m_{(x,y)}}$ are close together. However, when they are far apart, more weight will be placed on $Z_{m_{(x,y)}}$. The reasoning behind this is that typically monocular depth estimates capture quite well the rough structure of the scene, while stereo vision estimates are typically more accurate, but when wrong can result in quite large outliers. This leads to the following formula:
\begin{equation}
\label{eq:ratioWeight}
W_{s_{(x,y)}} =  \begin{cases} \frac{ N_{Z_{m}(x,y)} }{N_{Z_{s}(x,y)}} \enspace if \enspace N_{Z_{s}(x,y)} > N_{Z_{m}(x,y)}  \\
\frac{ N_{Z_{s}(x,y)} }{N_{Z_{m}(x,y)}} \enspace if \enspace N_{Z_{s}(x,y)} < N_{Z_{m}(x,y)}  \end{cases} 
\end{equation}
\noindent where $N_{Z_{m}(x,y)} = Z_{m_{(x,y)}} / \textrm{max}(Z_m)$ and $N_{Z_{s}(x,y)} = Z_{s_{(x,y)}} / \textrm{max}(Z_s)$. 

Finally, after the merging operation a median filter with a $5 \times 5$ kernel is used to smooth the final depth map and reduce even more overall noise. 

\section{Off-line experimental results}
\label{sec:experimentalResults}
To evaluate the performance of the merging algorithms the error metrics commonly found in the literature \cite{Eigen2014DepthMP} are used:

\begin{itemize}
\item Threshold error: \% of $y$ s.t. $max(\frac{y}{y^*}, \frac{y^*}{y}) = \delta < thr$
\vspace{7.5 pt}
\item Mean absolute relative difference: $\frac{1}{|N|} \sum_{y \in N} \frac{|y-y^*|}{y^*}$
\vspace{7.5 pt}
\item Mean squared relative difference: $\frac{1}{|N|} \sum_{y \in N} \frac{||y-y^*||^2}{y^*}$
\vspace{7.5 pt}
\item Mean linear RMSE: $\sqrt{\frac{1}{|N|} \sum_{y \in N}||y-y^*||^2}$
\vspace{7.5 pt}
\item Mean log RMSE:$\sqrt{\frac{1}{|N|} \sum_{y \in N}||\log y-\log y^*||^2}$
\vspace{7.5 pt}
\item Log scale invariant error:\\ $\frac{1}{2N} \sum_{y \in N} \left( \log y - \log y^* + \frac{1}{N}\sum_{y \in N}(\log y^* - \log y) \right)^2$ 
\end{itemize}
\noindent, where $y$ and $y^*$ are the estimated and corresponding ground truth depth in meters, respectively, and $N$ is the set of points. 


The main results of the experiments are summarized in Table \ref{table_errors}. Note that stereo vision is evaluated separately on non-occluded pixels with ground truth and on all pixels with ground-truth. The other estimators in the table are always applied to all ground-truth pixels. The results of the proposed fusion scheme are shown on the right in the table (FCN), and on the left the results are shown for three variants that all leave out one part of the merging algorithm. Surprisingly, a version of the merging algorithm without monocular scaling actually works the best, and also outperforms the stereo vision algorithm more clearly than the merging algorithm with scaling in Table \ref{table_errors}. Still, for what follows, we report on the fusion results with scaling.

\begin{table*}[t]
\centering
\caption{Errors per method. Best results on all ground-truth pixels in bold.}
\label{table_errors}
\begin{tabular}{l|ll|ll|ll|ll|ll}
                                 & \multicolumn{2}{c|}{\textbf{No weighting}} & \multicolumn{2}{c|}{\textbf{No monocular scaling}} & \multicolumn{2}{c|}{\textbf{Average scaling}} & \multicolumn{2}{c|}{\textbf{Stereo}} & \multicolumn{2}{c}{\textbf{FCN}} \\
                                 & SSL                & Fused SSL             & SSL                      & Fused SSL               & SSL                 & Fused SSL               & Non-occ only        & Incl Occ       & SSL             & Fused SSL           \\ \hline
threshold $\delta \textless 1.25$   & 0,38               & 0,52                  & 0,72                     & \textbf{0,85}           & 0,25                & 0,77                   & 0,92         & 0,88                          & 0,38            & 0,60                \\
threshold $\delta \textless 1.25^2$ & 0,68               & 0,82                  & 0,91                     & \textbf{0,96}           & 0,44                & 0,95                   & 0,97         & 0,92                          & 0,68            & 0,95                \\
threshold $\delta \textless 1.25^3$ & 0,84               & 0,93                  & 0,96                     & \textbf{0,98}           & 0,59                & 0,97                   & 0,98         & 0,93                          & 0,84            & 0,98                \\
abs relative difference          & 0,31               & 0,25                  & 0,20                     & \textbf{0,20}           & 0,48                & 0,23                   & 0,26         & 0,29                          & 0,31            & 0,24                \\
sqr relative difference          & 3,38               & 2,20                  & \textbf{2,16}            & 3,00                    & 4,98                & 3,50                   & 7,61         & 7,63                          & 3,38            & 3,17                \\
RMSE (linear)                    & 10,22              & 7,86                  & 7,67                     & \textbf{5,47}           & 10,24               & 5,92                   & 6,29         & 7,19                          & 10,22           & 6,14                \\
RMSE (log)                       & 0,48               & 0,36                  & 0,27                     & \textbf{0,24}           & 1,01                & 0,33                   & 0,28         & 2,46                          & 0,48            & 0,30                \\
RMSE (log, scale inv.)           & 0,05               & 0,04                  & 0,03                     & \textbf{0,03}           & 0,31                & 0,05                   & 0,04         & 2,95                          & 0,05            & 0,03               
\end{tabular}
\end{table*}

\begin{figure*}
    \centering
    \includegraphics[width=\textwidth]{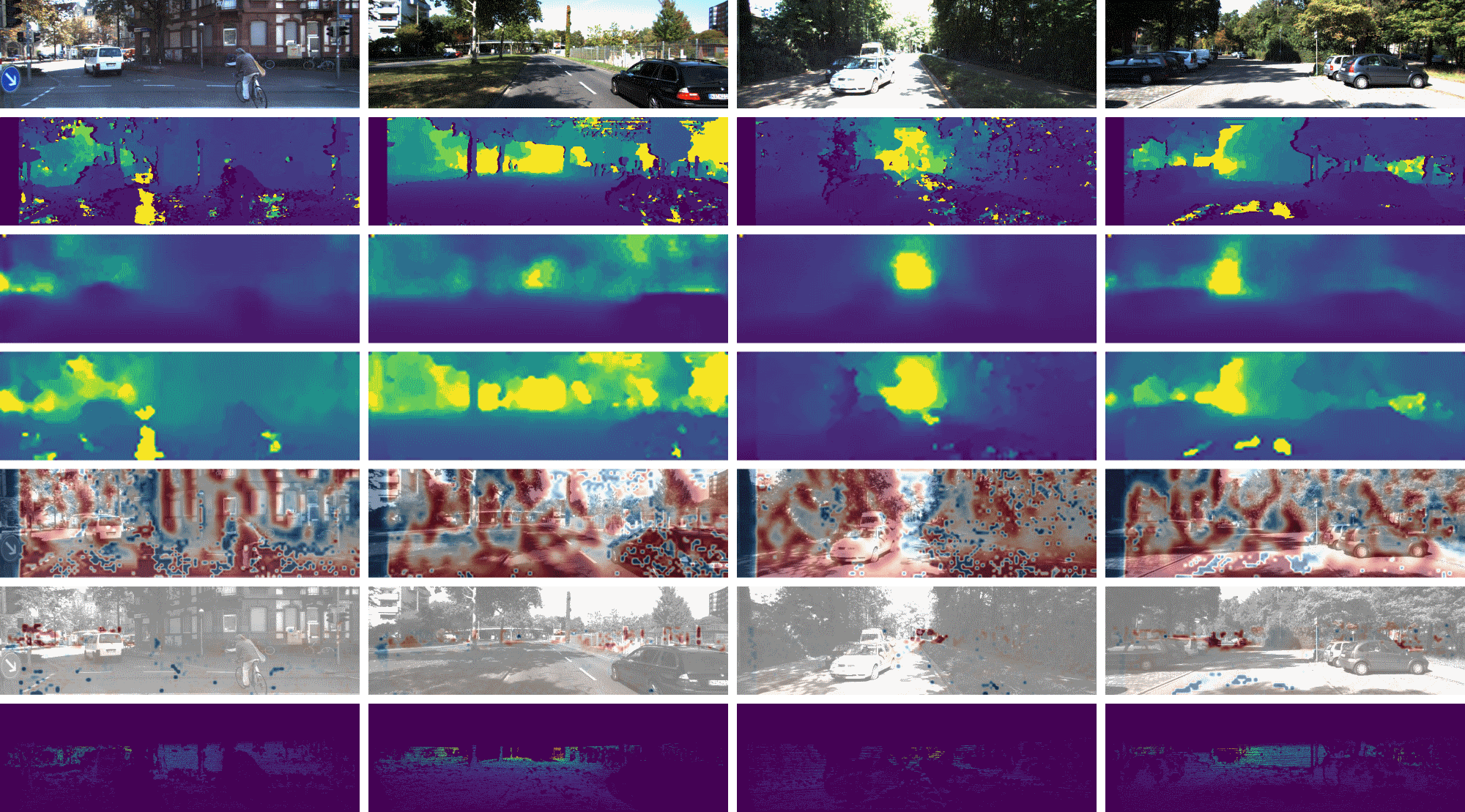}
\caption{Visual comparison between different depth maps using the same color scheme. Five images from Kitti were selected. Row 1) the rgb image. 2) Stereo depth map. 3) Still-mono depth map. 4) The merged depth map. 5) The confidence map (red is high stereo confidence, blue for mono). 6) The difference in error when compared against the Velodyne ground truth between mono and stereo (red for high mono errors, blue for high stereo errors). 7 Velodyne depth map. }
    \label{fig:comparison_a}
\end{figure*}
\begin{figure*}
    \centering
    \includegraphics[width=\textwidth]{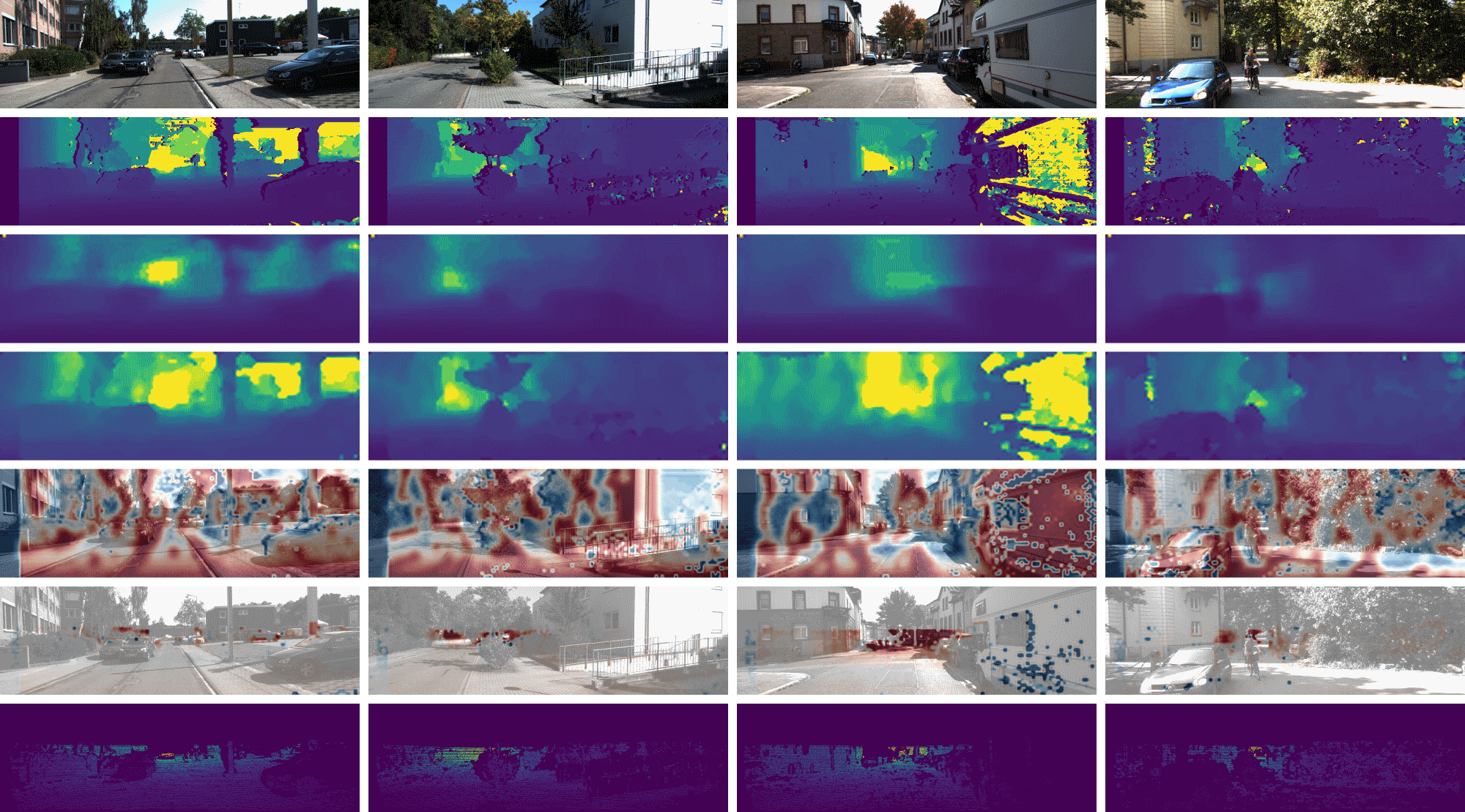}
\caption{Part 2, same legend as \ref{fig:comparison_a} }
    \label{fig:comparison_b}
\end{figure*}

\begin{figure}[h]
    \centering
    \includegraphics[scale =0.45]{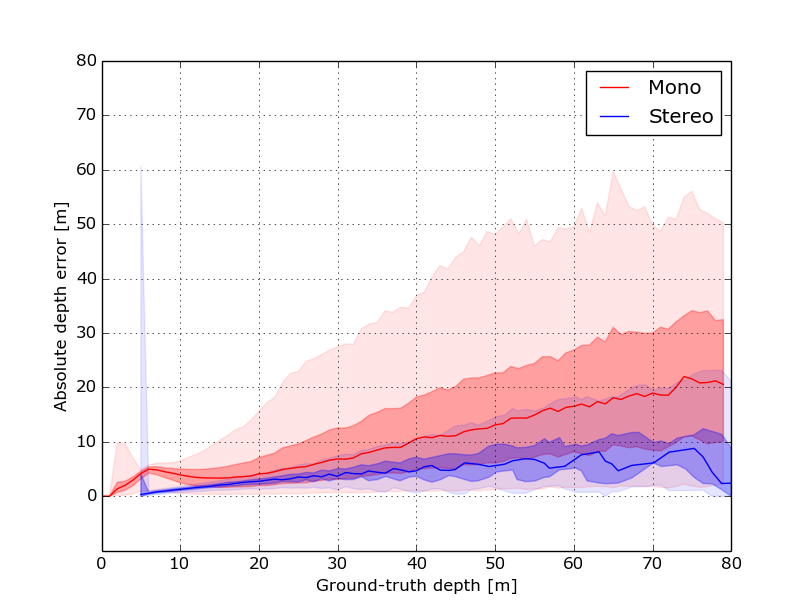}
\caption{Relation between the absolute error $|y-y^*|$ (y-axis) and the ground truth distance $y^*$ (x-axis). The solid lines indicate the median of the error distributions, while the light shading goes from $5\%$ to $95\%$ of the distribution and the darker shading from $25\%$ to $75\%$. The results for monocular depth estimation are shown in red, while the results for stereo depth estimation are shown in blue.}
    \label{fig:DVE}
\end{figure}

In order to get insight into the fusion process, we investigate the absolute errors of the stereo and monocular depth estimators as a function of the ground-truth depth obtained with the laser scanner. The results can be seen in fig. \ref{fig:DVE}. 

\begin{figure}[h]
    \centering
    \includegraphics[scale =0.45]{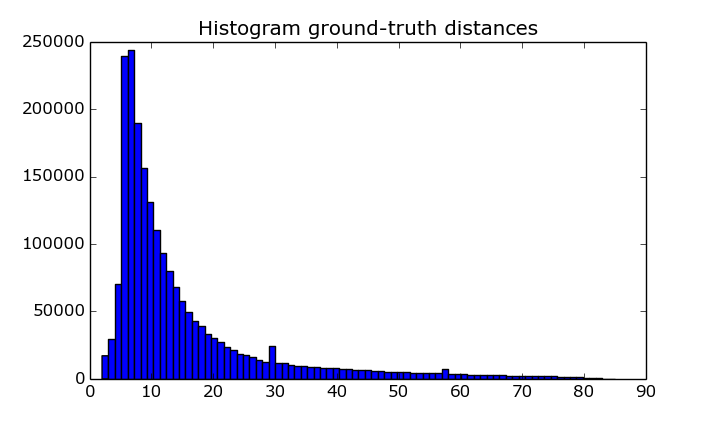}
\caption{Histogram of the depths measured by the Velodyne laser scanner.}
    \label{fig:hist_distances}
\end{figure}

We make five main observations. First, in comparison to the FCN monocular estimator, stereo vision in general gives more accurate depth estimates, also at the larger depths. Second, it can be seen that the monocular estimator provides depth values that are closer than stereo vision, which was limited to a maximal disparity of 64 pixels. Third, the accuracy of stereo vision becomes increasingly `wavy' towards 80 meters. This is due to the nature of stereo vision, in which the distance per additional pixel increases nonlinearly. The employed code determines subpixel disparity estimates up to a sixteenth of a pixel, but this does not fully prevent the increasing error when between pixel disparities further away. Fourth, stereo vision has a big absolute error peak at the low distances. This is due to large outliers, where stereo vision finds a better match at very large distances. Fifth, one may think that the monocular depth estimation far away is too bad for fusion. However, one has to realize that these results are made without scaling the monocular estimates - which can go beyond 80 meters, resulting in large errors. Moreover, investigation of the error $(y-y^*)$ shows that the monocular estimate is not biased in general. Finally, one has to realize that the majority of the pixels in the KITTI dataset lies close by, as can be seen in fig. \ref{fig:hist_distances}. Hence, the closer pixels are most important for the fusion result.

Figure \ref{fig:comparison_a} and \ref{fig:comparison_b} provide a qualitative assessment of the results. It illustrates three of the findings. First, the stereo vision depth maps show that often close-by objects are judged to be very far away (viz. the peak in fig. \ref{fig:DVE}. The most evident example is the Recreational Vehicle (RV) in the third column of fig. \ref{fig:comparison_b}. Second, the proposed fusion scheme is able to remove many of the stereo vision errors. Evidently, all occluded areas are filled in by monocular vision, improving depth estimation there. It also removes many of the small image patches mistakingly judged as very far away by the stereo vision. However, fusion does not always solve the entire problem - the aforementioned RV is an evident example of this. Indeed, the corresponding image in the fifth row shows that the fusion scheme puts a high weight on the stereo vision estimates (red), while the error in these regions is much lower for mono vision (blue in the sixth row). To help the reader interpret the fifth and sixth row; Ideally, if an image coordinate is colored in the sixth row (meaning that one method has a much higher error than the other), the image in the fifth row (confidence map) should have the same color. Third, the red areas in the images in the sixth row illustrate that monocular estimates are indeed less good than stereo vision at long distances.

\section{On-board experimental results}
\label{sec:onboard}
In order to investigate whether SSL stereo and mono fusion can also lead to better depth estimation on a computationally restricted robot, we have performed tests on board of a small drone. The experimental setup consisted of a Parrot SLAMDunk coupled to a Parrot Bebop drone and aligned with its longitudinal body axis. The stereo estimation used the semi-global block matching  algorithm \cite{Hirschmller2008StereoPB} also used in the off-board experiments. On board, we used the raw disparity maps without any type of post-processing. For monocular depth perception we used a very light-weight Convolutional Neural Network (CNN), i.e., only the coarse network of Ivanecky's CNN \cite{Thesis:Janivecky}. Due to computational limitations it was not possible to run the full network on-board. In these experiments, we use the network weight that were trained for the images of the NYU V2 data set and predicts depths up to 10 meters.

\begin{figure}[h]
    \centering
    \includegraphics[width = 150 pt, height =180 pt]{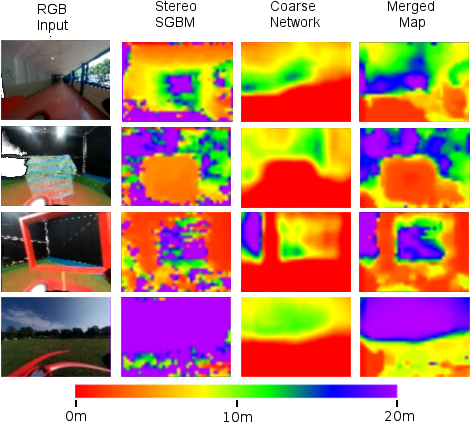}
\caption{\textit{From left to right:} left RGB input, stereo depth map, output of coarse network and final merged map. }
    \label{fig:flightResults}
\end{figure}

To test the performance of the merging algorithm the drone was flown both indoors and outdoors. The  algorithms ran in real-time at 4 frames per second with all the processing being done  on board. Selected frames and corresponding depth maps from the test flights are shown in fig. \ref{fig:flightResults}. 

There are clear differences between the three depth estimates. The stereo algorithm provides a semi-dense solution contaminated with a lot of noise (sparse purple estimates). Its performance is significantly deteriorated by the presence of the blades in lower regions of the images. The coarse network provides a solution without too much detail but where it is possible to understand the global composition of the scene. Finally, the merged depth map provides the most reliable solution. Except for the first row, where the bad monocular prediction induces errors in the final prediction, the merged map has more detail, less noise and the relative positions of the objects are better described. Although very preliminary, and in the absence of a ground-truth sensor, these results are promising for the on-board application of the proposed self-supervised fusion scheme.

\section{CONCLUSION}
\label{sec:conclusion}
In this article we investigated the fusion of a stereo vision depth estimator with a self-supervised learned monocular depth estimator. To this end, we presented a novel algorithm for dense depth fusion that preserves stereo estimates in high stereo confidence areas and uses the output of a CNN to correct for possibly wrong stereo estimates in low confidence and occluded regions. The experimental results show that the proposed self-supervised fusion indeed leads to better results. The analysis suggests that in our experiments, stereo vision is more accurate than monocular vision at most distances, except close by. 

We identify three main directions of future research. First, the current fusion of stereo and mono vision still involves a predetermined fusion scheme. This, while the accuracy of the monocular estimates may depend on the environment and hardware. For this reason, in \cite{GuidoC}, the robot used the trusted primary cue to determine the uncertainty of the learned secondary cue in an online process. This uncertainty was then used for fusing the two cues. A similar setup should be investigated here. Second, the performance obtained with our proposed fusion scheme is significantly lower than that of the image-reconstruction-based SSL setup in \cite{godard2017unsupervised}. We did not perform any thorough investigation of network structure, training procedure, etc. to optimize the monocular estimation performance, but such an effort would be of interest. Third, and foremost, we selected this task, as stereo vision is typically considered to be very reliable. The fact that even sub-optimal monocular estimators can be fused with stereo to improve the robot's depth estimation, is encouraging for finding other application areas of SSL fusion.



\end{document}